\documentclass[runningheads]{llncs}
\usepackage[T1]{fontenc}
\usepackage{graphicx}
\usepackage{diagbox}
\usepackage{xcolor}
\usepackage{makeidx} 
\usepackage[colorlinks,linkcolor=blue]{hyperref}
\usepackage{amsmath}
\usepackage[misc]{ifsym} 
\usepackage{bbding}
\usepackage{amsfonts}
\usepackage{xspace}
\usepackage{makecell}
\usepackage{enumitem} 
\usepackage{adjustbox}
\usepackage{tabularx}   
\usepackage{amssymb}    

\usepackage[compatibility=false]{caption}
\usepackage{booktabs}
\usepackage{multirow}
\usepackage{bm}      
\usepackage{amsbsy}
\usepackage{pdfrender}

\newcommand{\bfnum}[1]{%
    \leavevmode
    \rlap{#1}\kern0.02pt%
    \rlap{#1}\kern0.02pt%
    \rlap{#1}\kern0.02pt%
    #1%
}
\usepackage{graphicx,verbatim}
\begin{document}
%
\title{Rethinking Medical Landmark Localization with Prototype Learning-based Progressive \\Offset Correction}


\titlerunning{PPOC-LL}
%

\author{
Jingxian Xu\inst{1}\thanks{Jingxian Xu and Yuhao Huang contributed equally to this work.}\and 
Yuhao Huang\inst{1,2,3\star} \and
Rusi Chen\inst{1} \and
Yanfeng Zhou\inst{4} \and \\
Dong Ni\inst{1,4,5,6}\textsuperscript{(\Letter)}
}

\authorrunning{J. Xu and Y. Huang et al.}

\institute{
Medical Ultrasound Image Computing (MUSIC) Lab, Shenzhen University, Shenzhen, China\\
\email{nidong@szu.edu.cn}\\
\and
Centre for Artificial Intelligence and Robotics, Hong Kong Institute of Science \& Innovation, Chinese Academy of Sciences, Hong Kong, China 
\and
Boston Children's Hospital, Harvard Medical School, Boston, USA 
\and
School of Artificial Intelligence, Shenzhen University, Shenzhen, China
\and
School of Biomedical Engineering and Informatics, Nanjing Medical University, Nanjing, China
\and
National Engineering Laboratory for Big Data System Computing Technology, Shenzhen University, Shenzhen, China
}
  
\maketitle
\begin{abstract}

Accurate landmark localization in medical images is a fundamental step for quantitative clinical measurement and downstream analysis. Existing localization methods have advanced, among which multi-stage refinement is a superior solution. Although this strategy mitigates the anatomical ambiguity inherent in single-stage global predictions, its high computational cost limits practical applicability. In this work, we propose a parameter-economic model, PPOC-LL, which leverages  \textbf{P}rototype learning-based \textbf{P}rogressive \textbf{O}ffset \textbf{C}orrection for \textbf{L}andmark \textbf{L}ocalization. Our contribution is three-fold. \textbf{First}, to drive coarse-to-fine landmark optimization, we introduce a multi-scale dynamic perception strategy for patch-level feature pyramid modeling. \textbf{Second}, to effectively handle anatomically similar patterns, we design a similarity-driven prototype learning mechanism that captures informative local semantics for robust offset prediction. \textbf{Last}, to stabilize the model learning and improve the overall performance, we incorporate a novel error-aware reliability regularization via tolerance-based balancing. We collected a large validation cohort, including two public and one private datasets spanning X-ray and ultrasound modalities, covering cephalometric, symphysis-fetal head, and fetal heart landmarks. Extensive experiments demonstrate that PPOC-LL achieves satisfactory performance with a favorable trade-off between accuracy and model complexity.

\keywords{Landmark localization\and Multi-scale\and Prototype Learning.}

\end{abstract}

\section{Introduction}
Landmark localization constitutes a core element of quantitative measurements in diverse clinical applications~\cite{wang2016benchmark,bai2026iugc}, including craniofacial evaluation, intrapartum fetal monitoring, and fetal cardiac assessment, etc. 
However, manual annotation is time-consuming and labor-intensive. Moreover, anatomical ambiguity and inter-observer variability may introduce systematic bias, potentially compromising downstream measurements and clinical decision-making. Therefore, robust automated localization methods are urgently needed.

Recent deep learning-based studies have made significant progress in intelligent medical image analysis~\cite{hu2021joint,chen2020region,liang2022sketch,huang2024segment,huang2025flip}.
Specifically, for landmark localization, regression-based methods are one of the most common solutions, and can be broadly categorized into direct coordinate prediction and heatmap-based formulations.
The former predicts absolute landmark coordinates in an end-to-end manner~\cite{wu2017facial}, while the latter estimates per-landmark heatmaps and extracts coordinates from their peaks~\cite{chen2023automatic}. 
Zhou et al.~\cite{zhou2021learn} explored metric learning to strengthen the representation capability of heatmap-based landmark detectors.
Then, subsequent efforts integrated topology-aware constraint~\cite{huang2024landmark} and uncertainty quantification~\cite{jonkers2026reliable} were further leveraged to refine the quality and reliability of heatmap predictions.
However, most of them rely primarily on global information and perform prediction in a single forward pass, lacking explicit modeling of local contextual information around each landmark.

To address the above issues, several studies have investigated iterative approaches that incorporate local contextual information, with representative approaches including reinforcement learning (RL) and cascaded models.
Early RL typically learn a policy to sequentially navigate toward target landmarks within medical volumes~\cite{alansary2019evaluating}.
Alternatively, cascaded models first generate coarse predictions and subsequently refine them using local cues or higher-resolution representations.
Alignment-proposal-refinement joint learning~\cite{zeng2021cascaded} and discretized ordinary differential equation formulation~\cite{he2023cascade} have been specifically designed for cephalometric landmark localization.
Subsequently, Khan et al.~\cite{khan2024enhancing} extended the two-stage cascaded model to multi-resolution and multi-modal datasets.
In addition, a cascaded iterative Transformer was proposed for facial landmark detection under large pose and occlusion~\cite{li2024cascaded}.
However, RL-based methods may encounter training difficulties, and cascaded network designs typically increase model complexity, potentially limiting their applicability in clinical environments.

In this study, we propose the Prototype learning-based Progressive Offset Correction framework for Landmark Localization (named PPOC-LL).
Our highlights are as follows. 
First, we develop a multi-scale feature perception mechanism to extract feature pyramids and obtain initial landmark predictions.
These cues facilitate coarse-to-fine attention modeling and enable patch-level progressive refinement.
Second, we introduce a prototype learning scheme combined with local matching to capture vital semantic representations, enabling robust offset estimation.
Last, we design a tolerance-guided regularization that ensures stable model learning and improved detection accuracy.
Experiments on three medical datasets spanning two imaging modalities show that PPOC-LL outperforms state-of-the-art methods.
Moreover, PPOC-LL effectively balances performance and model complexity, indicating strong potential for clinical applicability.

\section{Method}
\begin{figure}[!t]
    \centering
    \includegraphics[width=1.0\textwidth]{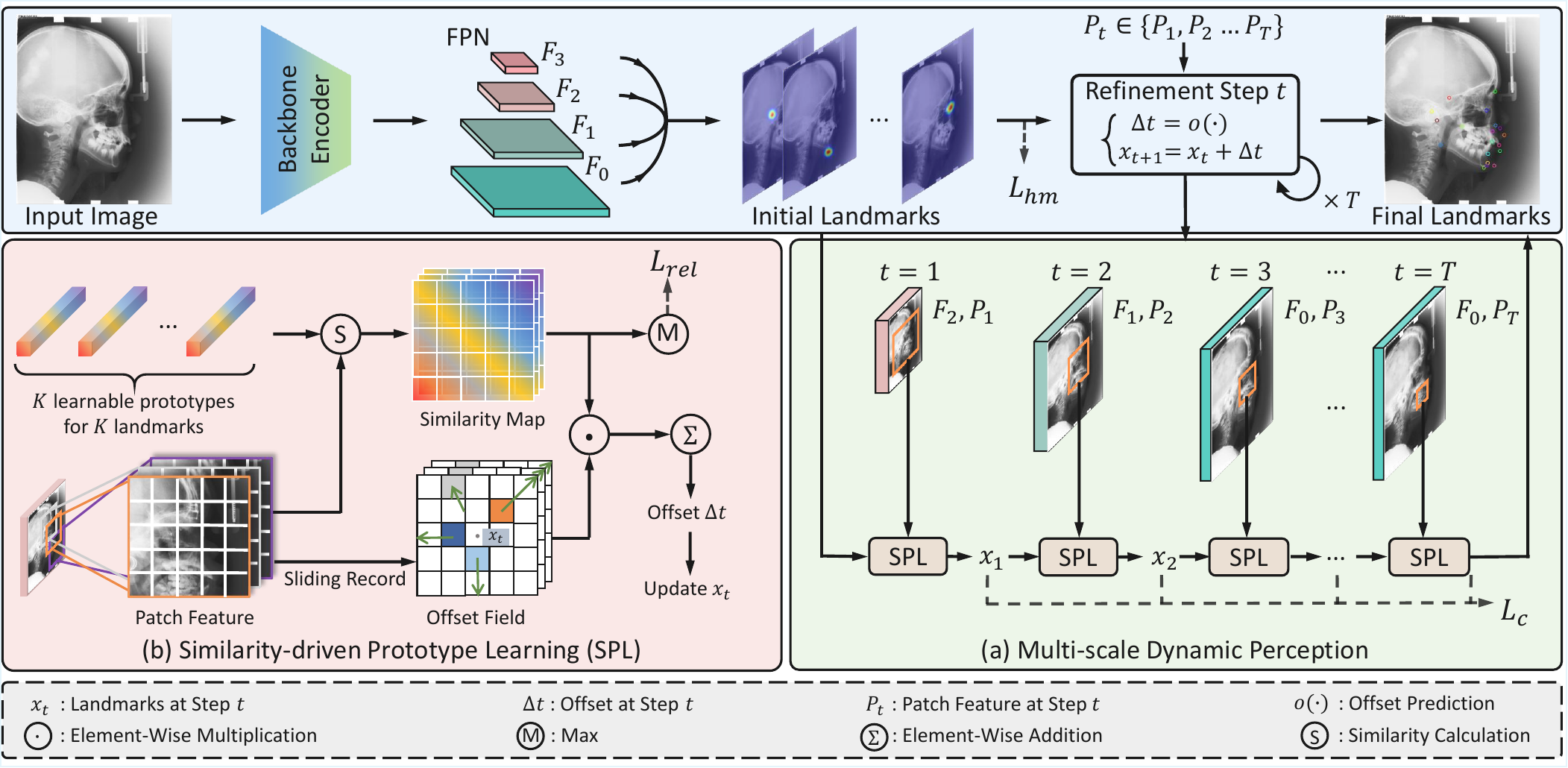}
    \caption{Illustration of our proposed PPOC-LL.}
    \label{fig:method}
\end{figure}

Fig.~\ref{fig:method} shows our PPOC-LL framework, which treats landmark localization as progressive correction via offset prediction.
First, a regular backbone equipped with a feature pyramid network (FPN) outputs multi-scale knowledge and initial landmarks via \textit{soft-argmax}.
Then, the proposed multi-scale patch-based dynamic perception and similarity-driven prototype learning strategies capture the offset relationships and update the landmark position progressivly.
Besides, an error-aware reliability loss is equipped to stabilize the training process. 

\subsection{Multi-scale Dynamic Perception for Progressive Optimization}

Existing single-pass landmark localization relied on global context and struggled to resolve local anatomical similarity.
Although FPN provides multi-scale representations, a fixed-scale sampling strategy is suboptimal for progressive refinement, since early corrections require a larger search range while later updates benefit from finer spatial details.
Therefore, we build PPOC-LL on top of an FPN with feature maps $F_0$-$F_3$, and perform iterative refinement by sampling multi-scale patches in a coarse-to-fine schedule.
Specifically, each refinement step $t$ samples a compact $p_t \times p_t$ feature patch centered at the current estimate $x_t$ on the selected pyramid scale to decode a displacement for location update.

In our setting, early steps rely on lower-resolution features with larger receptive fields for coarse correction, whereas later steps refine landmarks using higher-resolution features for improved precision.  
The patch is always centered at the $x_t$, yielding step-specific updates that form a continuous correction trajectory.
This step-wise re-centering of the sampled patch around $x_t$ is termed as a \textit{sliding record}.
We define the perception scale at step $t$ using an input image space search radius $R_{t}$, and map it to a selected pyramid level $l_t$ from $\{F_0,F_1,F_2\}$.
The conversion factor between the input and the $l_t$-th feature is:
\begin{equation}
s_{l_t} = (W - 1)/(W_{l_t} - 1),
\end{equation}
where $W$ and $W_{l_t}$ denote spatial resolutions of input image and the $l_t$-th pyramid, respectively. 
With the $ceiling$ function, the patch size can be formulated as:
\begin{equation}
p_t = 2 \left\lceil R_t/s_{l_t} \right\rceil + 1.
\end{equation}

\subsection{Similarity-driven Prototype Learning for Offset Correction}
\label{sec:2.2}
In iterative refinement, each update is driven by evidence collected around the current landmark location. However, relying solely on local responses can be unstable under appearance variations across subjects and acquisition conditions, where the same landmark may exhibit diverse textures and different landmarks may appear similar. 
To enhance discriminability, we associate each landmark with a learnable prototype vector as a semantic reference.

Specifically, at step $t$, the local patch feature for landmark $k$ is defined as ${P_{t,k}}$, centered on ${x_{t,k}}$. 
We then project the ${P_{t,k}}$ to the prototype dimension using 1$\times$1 convolution ($conv$) and compute a cosine similarity $M_{t,k}(i,j)$ map between the learnable prototype $q_{k}$ and each spatial location $(i, j)$ in the patch, as follows:
\begin{equation}
M_{t,k}(i,j) = cos(q_{k}, conv(P_{t,k}(i, j))).
\end{equation}

We then compute a normalized local matching distribution $m_{t,k}(i,j)$ by applying a softmax over the similarity map $M_{t,k}(i,j)$ across spatial locations in the patch.
Finally, we decode the displacement as the expectation over a predefined offset field $\Delta(i,j)$ on the sampled patch, and update the landmark locations:
\begin{equation}
\begin{aligned}
& \Delta_{t,k} = \sum_{i,j} \ m_{t,k}(i,j) \Delta(i,j), \\
& \qquad x_{t+1,k} = x_{t,k} + \Delta_{t,k},
\end{aligned}
\end{equation}
where $\Delta(i, j)=(\Delta x, \Delta y)$ is fixed and computed from the patch's relative indices. 
We set $\Delta x = (j - \lfloor p_t/2 \rfloor) \cdot 2 / (W_{l_t} - 1)$ and $\Delta y = (i - \lfloor p_t/2 \rfloor) \cdot 2 / (H_{l_t} - 1)$, where $(H_{l_t}, W_{l_t})$ is the spatial resolution of the feature map selected at step $t$, so that $\Delta(i, j)$ is expressed in the same normalized coordinate system as $x_{t, k}$.

\subsection{Error-aware Reliability Regularization for Stable Training}
We observe that the sharpness of the local matching distribution is not always calibrated to the underlying localization error during iterative refinement.
Therefore, coordinate-only supervision can not explicitly enforce consistency between the distribution confidence and the actual localization error, which can destabilize training.
To address this issue, we propose an error-aware reliability regularizer to provide a step-wise scalar reliability signal.
To be specific, the peak probability of local matching distribution $m_{t,k}$ serves as the predicted reliability.
\begin{equation}
r_{t,k} = \max_{i,j} \ m_{t,k}(i,j) \in [0, 1].
\end{equation}
Then, we derive the soft supervision target $\tilde{r}_{t,k}$ from the Euclidean localization error between the current estimate and the ground truth $x_k^*$:
\begin{equation}
\tilde{r}_{t,k} = 1 - \text{clip}\!\left( 
\frac{\left\| x_{t,k} - x_k^* \right\|_2}{\gamma},\, 0,\, 1 \right).
\end{equation}

When constructing the target, we detach $x_{t,k}$ so that this regularizer supervises the reliability prediction only and does not backpropagate to the coordinate updates.
The $\text{clip}(\cdot,0,1)$ operator clamps the normalized error to $[0,1]$, ensuring $\tilde{r}_{t,k}\in[0,1]$, where $\gamma$ is the tolerance parameter specifically defined by different datasets to ensure geometric consistency under multi-scale resizing.
This yields a monotonic and continuous relationship between localization error and reliability, casting the supervision as error regression rather than a hard threshold classification. Finally, the reliability regularization term can be defined by:
\begin{equation}
\mathcal{L}_{\text{rel}} =  \sum_{t=1}^{T} \sum_{k=1}^{K} |r_{t,k} - \tilde{r}_{t,k}|.
\end{equation}

We train the model with three losses: 
(i) a typical heatmap regression loss $\mathcal{L}_{hm}$ based on mean squared error, where the target heatmaps are generated from the ground-truth coordinates; 
(ii) a step-wise coordinate loss $\mathcal{L}_{c}$ that applying L1 penalties to each $x_{t}$ with larger weights on later steps; and 
(iii) the above-mentioned tolerance-balanced regularization $L_{rel}$. 
Finally, the total loss $\mathcal{L}$ is:
\begin{equation}
\mathcal{L} = \lambda_1 \mathcal{L}_{hm} + \lambda_2 \mathcal{L}_{c} + \lambda_3 \mathcal{L}_{rel}.
\end{equation}

\section{Experiments}
\textbf{Datasets and Evaluation Metrics.} 
We collected two public datasets (Cephalograms~\cite{wang2016benchmark}, CE and  Intrapartum Ultrasound Grand Challenge, IUGC~\cite{bai2026iugc}) and one private dataset (fetal heart ultrasound, FHU) for method validation. 
\textit{CE} comprises 400 lateral cephalogram images, each annotated with 19 landmarks. 
The dataset was officially split into 150, 150 and 100 images for training, test1 and test2, respectively.
\textit{IUGC} includes 801 mid-sagittal view images, with 3 landmarks on the pubic symphysis and fetal head annotated for Angle of Progression (AoP) calculation. 
Following the fully-supervised setting, we use 300 and 501 images for training and testing.
\textit{FHU} contains 905 mid-gestation four-chamber planes annotated with 24 anatomical landmarks, of which 724 are used for training and 181 for testing.
For \textit{CE} and \textit{FHU}, Mean Radial Error (MRE) and Successful Detection Rate (SDR) were selected as the evaluation metrics. 
The SDR thresholds were set to 2, 2.5, 3, and 4 mm for \textit{CE}, and 3, 4, 6, and 8 pixels for \textit{FHU}.
For \textit{IUGC}, we follow~\cite{bai2026iugc} to report MRE and also landmark-level statistics ($PS_R$/$PS_L$/$FH_t$), with AoP error for clinical measurement evaluation.

\begin{table}[!t]
\centering
\caption{Quantitative comparison on CE, FHU, and IUGC datasets. The best results are shown in \textbf{bolded}, and the second-best ones are \underline{underlined}.}
\label{tab:combined_results}
{\scriptsize
\begin{tabular*}{\textwidth}{@{\extracolsep{\fill}}ccccccccccc}
\toprule
\multicolumn{11}{c}{\text{CE}}\\
\midrule
\multirow{3}{*}{Methods} & \multicolumn{5}{c}{Test1} & \multicolumn{5}{c}{Test2} \\
\cmidrule(l{2pt}r{0pt}){2-6} \cmidrule(l{2pt}r{0pt}){7-11}
 &  \multirow{2}{*}{MRE$\downarrow$} & \multicolumn{4}{c}{SDR$\uparrow$} & \multirow{2}{*}{MRE$\downarrow$} & \multicolumn{4}{c}{SDR$\uparrow$} \\
\cmidrule(l{2pt}r{0pt}){3-6} \cmidrule(l{2pt}r{0pt}){8-11}
 & & 2mm & 2.5mm & 3mm & 4mm & & 2mm & 2.5mm & 3mm & 4mm \\
\midrule

CenterNet~\cite{zhou2019objects} & 2.13$_{1.41}$ & 55.43 & 70.32 & 80.57 & 90.49 & 2.49$_{1.95}$ & 47.68 & 62.53 & 73.72 & 84.42 \\
Multi-Reg~\cite{lee2022cephalometric} & 1.19$_{\text{\bfnum{0.80}}}$ & 86.42 & 92.00 & 95.54 & \underline{98.53} & - & 74.58 & 81.79 & 87.53 & 94.26 \\
APR~\cite{zeng2021cascaded} & 1.34$_{0.92}$ & 81.37 & 89.09 & 93.79 & 97.86 & 1.64$_{\underline{0.91}}$ & 70.58 & 79.53 & 86.05 & 93.32 \\
ContextNet~\cite{oh2020deep}& 1.18$_{1.01}$ & 86.20 & 91.20 & 94.40 & 97.70 & \underline{1.46}$_{\text{\bfnum{0.82}}}$ & \bfnum{75.90} & 83.40 & \underline{89.30} & 94.70 \\
AFPF-RV~\cite{chen2019cephalometric}& 1.17 & 86.67 & \underline{92.67} & 95.54 & \underline{98.53} & 1.48 & 75.05 & 82.84 & 88.53 & 95.05 \\
CH-Net~\cite{mccouat2022contour}& 1.20 & 83.47 & 89.16 & 92.60 & 96.49 & \underline{1.46} & 74.63 & \underline{83.58} & 87.21 & 93.79 \\
NFDP~\cite{huang2024landmark}& \underline{1.14}$_{0.93}$ & \underline{87.02} & 92.38 & \underline{95.76} & 98.35 & \underline{1.46}$_{1.31}$ & 75.11 & 82.48 & 89.16 & \underline{95.16} \\
\bfnum{PPOC-LL} & \bfnum{1.09}$_{\text{\underline{0.87}}}$ & \bfnum{88.49} & \bfnum{93.51} & \bfnum{95.89} & \bfnum{98.53} & \bfnum{1.43}$_{1.25}$ & \underline{75.79} & \bfnum{83.74} & \bfnum{89.42} & \bfnum{95.53} \\
\bottomrule
\end{tabular*}
}

{\scriptsize
\resizebox{\textwidth}{!}{%
\begin{tabular}{cccccc c cc ccccc}
\cmidrule[\heavyrulewidth]{1-6} \cmidrule[\heavyrulewidth]{8-14}
\multicolumn{6}{c}{FHU} & & \multicolumn{7}{c}{IUGC} \\
\cmidrule{1-6} \cmidrule{8-14}
\multirow{2}{*}{Methods} & \multirow{2}{*}{MRE$\downarrow$} & \multicolumn{4}{c}{SDR$\uparrow$} & &
\multirow{2}{*}{Strategy} & \multirow{2}{*}{Methods} & \multirow{2}{*}{MRE$\downarrow$} & \multirow{2}{*}{MRE} & \multirow{2}{*}{MRE} & \multirow{2}{*}{MRE} & \multirow{2}{*}{$\Delta$AoP} \\
\cmidrule{3-6}
& & 3px & 4px & 6px & 8px & & & & & ($PS_R$) & ($PS_L$) & ($FH_T$) &($^\circ$) \\
\cmidrule{1-6} \cmidrule{8-14}
IntegralNet~\cite{sun2018integral}& 9.73$_{8.86}$ & 11.35 & 18.14 & 34.28 & 49.70 & &
\multirow{3}{*}{\shortstack{Semi-\\supervised}} & T1~\cite{liu2025noisy} & 13.16$_{6.25}$ & 8.56$_{5.01}$ & 9.19$_{6.77}$ & 21.73$_{16.05}$ & 4.42$_{4.79}$ \\
DSNT~\cite{nibali2018numerical}& 8.92$_{\text{\underline{8.25}}}$ & 15.12 & 23.45 & 40.80 & 55.90 & &
&T2~\cite{ma2025unlabeled} & 11.67$_{6.34}$ & 6.53$_{4.38}$ & 8.60$_{5.06}$ & 19.90$_{17.54}$ & 3.81$_{3.12}$ \\
HRNet~\cite{wang2020deep}& 9.95$_{9.15}$ & 8.75 & 15.65 & 30.57 & 47.01 & &
&T8~\cite{deng2025two} & 12.82$_{6.14}$ & 7.81$_{5.15}$ & 9.25$_{7.64}$ & 21.42$_{15.88}$ & 4.57$_{7.90}$ \\
\noalign{\vskip -0.5pt}
\cline{8-14}
\noalign{\vskip 0.5pt}
SHG~\cite{newell2016stacked}& 10.30$_{9.62}$ & 8.66 & 14.71 & 30.02 & 45.86 & &
\multirow{3}{*}{\shortstack{Fully-\\supervised}} & T4~\cite{yang2025dsnt} & \underline{14.77}$_{\text{\bfnum{5.91}}}$ & \underline{9.06}$_{\text{\underline{5.70}}}$ & \underline{11.57}$_{\text{\bfnum{7.53}}}$ & \underline{23.69}$_{\text{\bfnum{14.83}}}$ & \underline{4.70}$_{\text{\underline{5.26}}}$ \\
NFDP~\cite{huang2024landmark} & \underline{7.57}$_{\text{\bfnum{6.62}}}$ & \bfnum{22.19} & \underline{32.83} & \underline{51.93} & \underline{66.85} & &
&T10~\cite{tang2025heatmap} & 21.83$_{19.89}$ & 10.67$_{9.27}$ & 15.62$_{25.74}$ & 39.19$_{51.87}$ & 8.37$_{14.64}$ \\
\bfnum{PPOC-LL} & \bfnum{7.34}$_{\text{\bfnum{6.62}}}$ & \underline{22.18} & \bfnum{34.12} & \bfnum{53.91} & \bfnum{67.70} & &
& \bfnum{PPOC-LL} & \bfnum{12.28}$_{\text{\underline{12.30}}}$ & \bfnum{8.12}$_{\text{\bfnum{5.46}}}$ & \bfnum{9.49}$_{\text{\underline{9.54}}}$ & \bfnum{19.23}$_{\text{\underline{16.18}}}$ & \bfnum{4.33}$_{\text{\bfnum{4.54}}}$ \\
\cmidrule[\heavyrulewidth]{1-6} \cmidrule[\heavyrulewidth]{8-14}
\end{tabular}%
}
}
\end{table}

\begin{figure}[!t]
    \centering
    \includegraphics[width=1.0\textwidth]{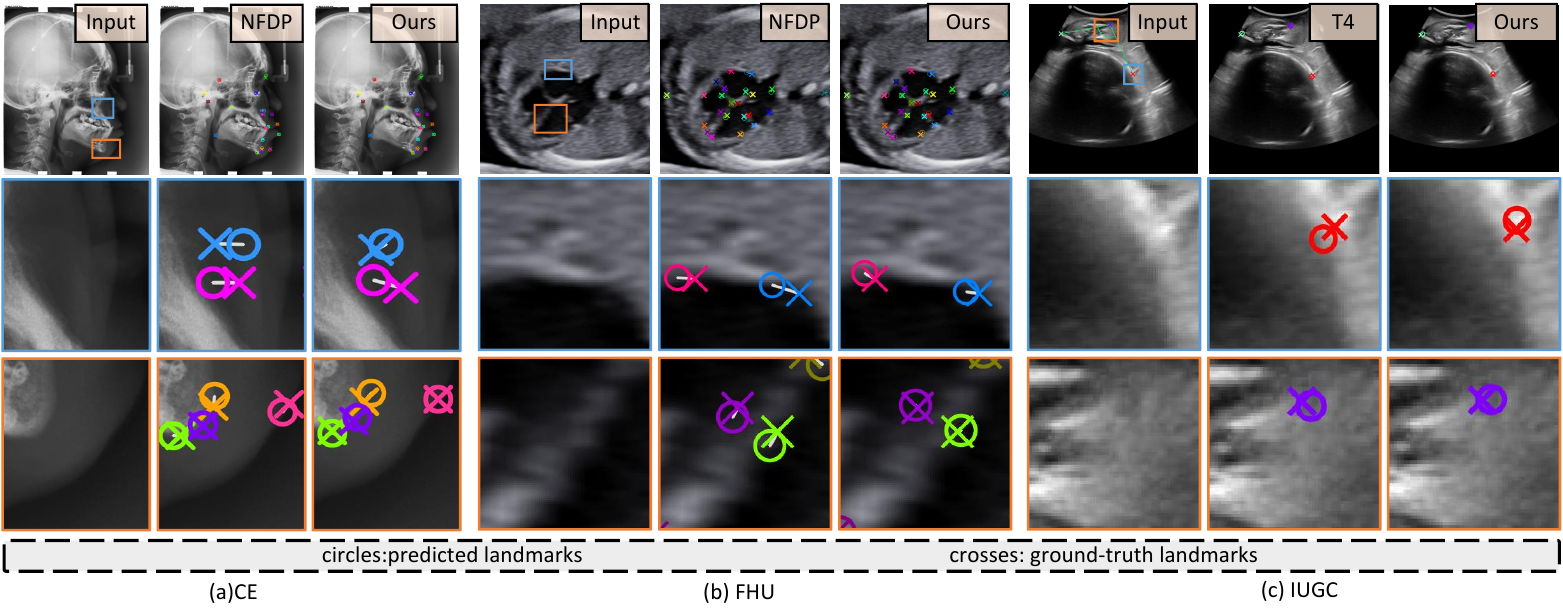}
    \caption{
    Qualitative results on CE, FHU, and IUGC datasets. Rows 2-3 show zoomed-in patches (see Row 1) for clearer landmark comparison.}  
    \label{fig:vis_all}
\end{figure}

\begin{table}[!t]
  \centering
  \begin{minipage}{0.48\textwidth}
    \centering
    \scriptsize
    \caption{Effectiveness of components on CE dataset, including Refinement Paradigm (RP) and Error-aware Regularization (ER). The best and second-best results are \textbf{bolded} and \underline{underlined}, respectively.}
    \label{tab:ablation_complete}
    \begin{tabular}{ccccccc}
      \toprule
      \multirow{2}{*}{$RP$} & \multirow{2}{*}{$ER$} & \multirow{2}{*}{MRE$\downarrow$} & \multicolumn{4}{c}{SDR $\uparrow$} \\
      \cmidrule(l{0pt}r{0pt}){4-7}
      & & & $2.0$mm & $2.5$mm & $3.0$mm & $4.0$mm \\
      \midrule
      & & $1.37$ & $78.48$ & $86.04$ & $91.05$ & $96.46$ \\
      \checkmark & & $1.31$ & \underline{80.38} & \underline{87.50} & \underline{91.91} & \underline{97.06} \\
      & \checkmark & \underline{1.30} & 80.07 & 87.45 & 91.69 & 96.67 \\
      \checkmark & \checkmark & \bfnum{1.26} & \bfnum{82.14} & \bfnum{88.60} & \bfnum{92.78} & \bfnum{97.12} \\
      \bottomrule
    \end{tabular}
  \end{minipage}\hfill
  \begin{minipage}{0.48\textwidth}
    \centering
    \includegraphics[width=0.8\linewidth]{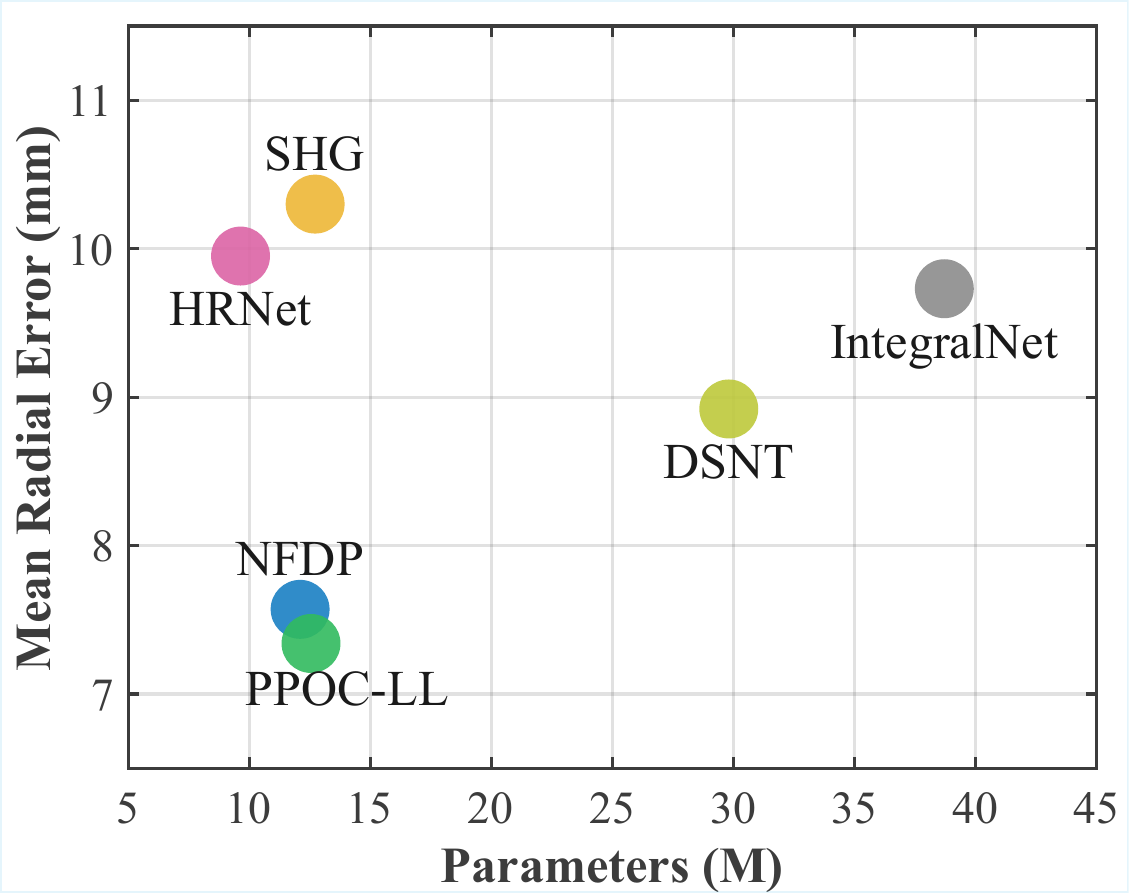}
    \captionof{figure}{Comparison of different methods in terms of MRE and model size.}
    \label{fig:efficiency_bubble}
  \end{minipage}
\end{table}

\begin{table}[!t]
  \centering
  \caption{Comparison of different iteration step $T$ and tolerance parameter $\gamma$ on CE-Test1 dataset. Best results are \textbf{bolded}, second-best are \underline{underlined}.}
  \label{tab:ablations_combined}
  
  \begin{minipage}[t]{0.48\textwidth}
    \centering
    \scriptsize
    \begin{tabular*}{\linewidth}{@{\extracolsep{\fill}}cccccc}
      \toprule
      \multirow{2}{*}{$T$} & \multirow{2}{*}{MRE$\downarrow$} & \multicolumn{4}{c}{SDR $\uparrow$} \\
      \cmidrule(l{0pt}r{0pt}){3-6}
      && 2.0mm & 2.5mm & 3.0mm & 4.0mm\\
      \midrule
      1 & 1.47$_{0.96}$ & 76.95 & 87.19 & 93.23 & 97.93 \\
      3 & 1.15$_{\underline{0.93}}$ & 86.35 & 92.42 & 95.54 & 98.42 \\
      4 & \underline{1.10}$_{\text{\bfnum{0.87}}}$ & 88.04 & \underline{93.12} & 95.82 & \underline{98.53} \\
      5 & \bfnum{1.09}$_{\text{\bfnum{0.87}}}$ & \bfnum{88.49} & \bfnum{93.51} & \underline{95.89} & \underline{98.53} \\
      6 & \underline{1.10}$_{\text{\bfnum{0.87}}}$ & \underline{88.18} & 92.98 & \bfnum{96.00} & \bfnum{98.60} \\
      \bottomrule
    \end{tabular*}
  \end{minipage}\hfill
  \begin{minipage}[t]{0.48\textwidth}
    \centering
    \scriptsize
    \begin{tabular*}{\linewidth}{@{\extracolsep{\fill}}cccccc}
      \toprule
      \multirow{2}{*}{$\gamma$} & \multirow{2}{*}{MRE$\downarrow$} & \multicolumn{4}{c}{SDR $\uparrow$} \\
      \cmidrule(l{0pt}r{0pt}){3-6}
      && 2.0mm & 2.5mm & 3.0mm & 4.0mm\\
      \midrule
      2  & 1.09$_{\text{\underline{0.88}}}$ & 87.72 & \underline{93.16} & 95.75 & \underline{98.53} \\
      3  & \bfnum{1.09}$_{\text{\bfnum{0.87}}}$ & \bfnum{88.49} & \bfnum{93.51} & \underline{95.89} & \underline{98.53} \\
      5  & \underline{1.11}$_{\text{\underline{0.88}}}$ & 87.47 & 92.60 & \bfnum{96.18} & 98.50 \\
      7  & 1.13$_{0.91}$ & \underline{87.86} & 92.84 & 95.86 & \bfnum{98.74} \\
      10 & 1.16$_{\text{\underline{0.88}}}$ & 86.77 & 92.53 & 95.47 & 97.96 \\
      \bottomrule
    \end{tabular*}
  \end{minipage}
\end{table}

\textbf{Implementation Details.}
We implement our model in PyTorch, using a single NVIDIA RTX 4090 GPU. ResNet18 is used as the backbone for all models. 
AdamW was used to optimize model training with a weight decay of 1e-4, a batch size of 16, and 400 epochs. A cosine annealing learning rate schedule is adopted with an initial learning rate of 2e-4. Data augmentation includes flipping, rotation, and transposition.
Input images are resized to $512 \times 512$ for \textit{CE} and \textit{IUGC}, and to $256 \times 256$ for \textit{FHU}.
We set the input image space search radius $R_t$ to $\{112, 80, 48, 48, 28\}$ for $512 \times 512$ inputs, and to $\{119, 56, 20, 20, 12, 8\}$ for $256 \times 256$ inputs.
We use $\lambda_1=1$, $\lambda_2=20$, and $\lambda_3=0.2$ for the loss weights.
Based on validation performance, we set the refinement steps $T$=$5,6,6$ and tolerance parameter $\gamma$=$3,3,8$ for \textit{CE}, \textit{IUGC}, and \textit{FHU}, respectively.
For PPOC-LL and ablation variants, experiments were repeated with three random seeds, and the averaged results are reported.
For MRE, the small values shown in the right-down corner indicate the standard deviation of radial errors on the test set.

\textbf{Comparisons with State-of-the-art Models.}
Table~\ref{tab:combined_results} compares PPOC-LL with various strong methods. 
Specifically, for CE and FHU, we included 7 and 5 competitors, respectively.
For IUGC, we report results in both semi-supervised (T1/T2/T8) and fully-supervised (T4/T10) settings~\cite{bai2026iugc}.
Results on CE and FHU indicate that PPOC-LL consistently outperforms most methods, ranking first or second across the evaluated metrics.
Specifically, PPOC-LL reports MREs of 1.09 mm, 1.43 mm, and 7.34 px on CE-Test1, CE-Test2, and FHU, respectively, surpassing the performance of the state-of-the-art NFDP.
Across the two CE test sets, PPOC-LL exhibits lower standard deviations (0.87/1.25) than NFDP (0.93/1.31), demonstrating superior cross-sample robustness.
On IUGC, PPOC-LL ranks first among fully supervised methods and, despite not using any unlabeled data, achieves performance comparable to semi-supervised approaches.
These gains strongly demonstrate the effectiveness of our overall design, including prototype-guided local matching and progressive offset correction.
Fig.~\ref{fig:vis_all} visualizes that the predictions from our PPOC-LL method are closer to the ground truth points.
Fig.~\ref{fig:efficiency_bubble} illustrates the trade-off between MRE and model size. It shows that our PPOC-LL obtains the lowest MRE among all methods, specifically outperforming NFDP with only a slight increase in parameters.

\begin{figure}[!t]
    \centering
    \includegraphics[width=1.0\textwidth]{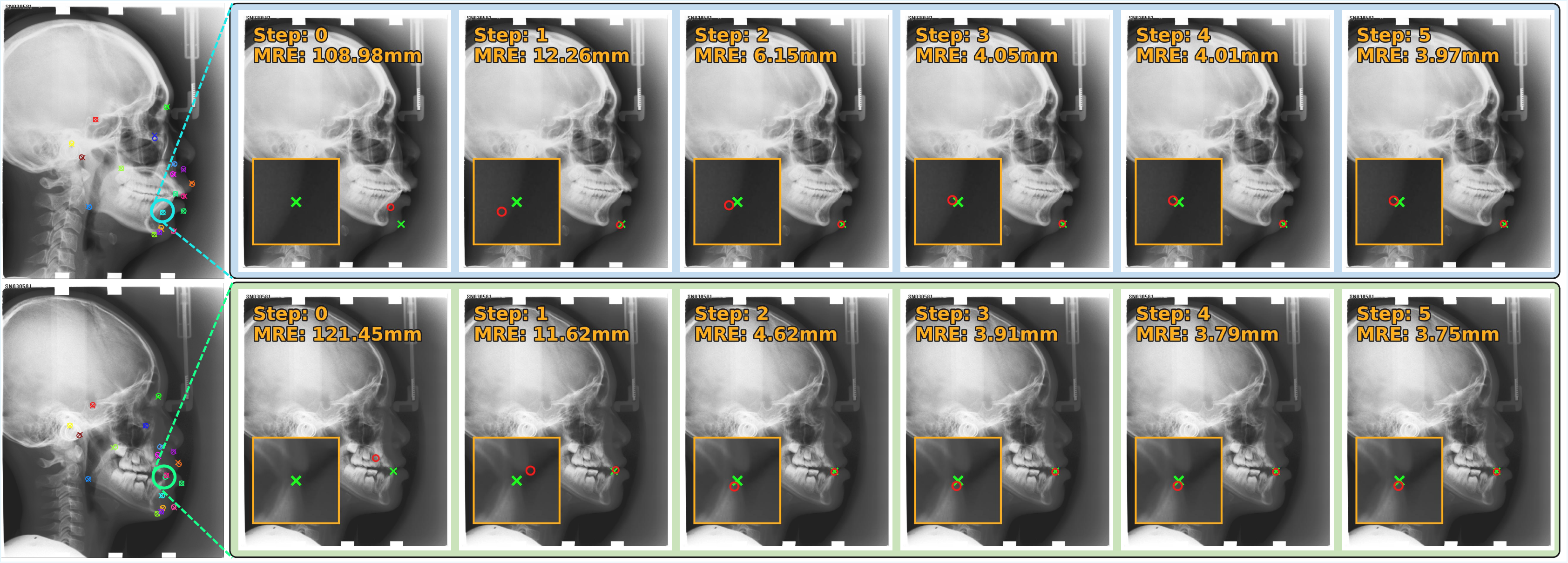}
    \caption{Visualization of the refinement process on the CE dataset. The lower-left zoomed-in panels show the progressive convergence of predicted landmarks (\textcolor{red}{circles}) toward the ground truths (\textcolor{green}{crosses}) over iterations.}
    \label{fig:ce_iter}
\end{figure}

\textbf{Ablation Studies.} 
As shown in Table~\ref{tab:ablation_complete}, compared with directly predicting landmarks in a single pass on the finest pyramid feature, introducing the progressive refinement paradigm can reduce the MRE from 1.37~mm to 1.31~mm and improve SDR across all thresholds, with the 2.0~mm SDR increasing from 78.48\% to 80.38\%. 
Incorporating the error-aware regularization will also decrease the MRE to 1.30~mm with consistent SDR gains. 
Equipped with both components can bring further improvement, achieving the lowest MRE of 1.26~mm and the highest SDR (82.14\% at 2.0~mm, 88.60\% at 2.5~mm, 92.78\% at 3.0~mm, and 97.12\% at 4.0~mm). 
These results suggest that the proposed strategies provide complementary benefits, leading to precise and stable landmark localization.

Table~\ref{tab:ablations_combined} further evaluates the impact of the iteration steps $T$ and the tolerance parameter $\gamma$ on model performance using the CE dataset.
Results show that increasing $T$ substantially improves both MRE ($\sim$0.38$\downarrow$) and SDR ($\sim$12$\uparrow$, 2.0 mm), with the performance gradually plateauing around $T=5/6$.
We also present cephalometric examples in Fig.~\ref{fig:ce_iter} to show the refinement process.
These results demonstrate the effectiveness of our designed multi-scale iterative refinement and suggest an optimal number of iteration steps.
For different $\gamma$, PPOC-LL exhibits high robustness, with MRE remaining stable within a narrow range (1.09-1.16 mm) as $\gamma$ varies from 2 to 10.
Such insensitivity shows that our regularization strategy can stabilize training without extensive hyperparameter tuning.

\section{Conclusion}
In this work, we propose a novel framework, named PPOC-LL, for automated landmark localization that integrates progressive offset correction with prototype learning across three medical datasets. 
Specifically, we first design a patch-level dynamic perception strategy to exploit multi-scale feature pyramids and iteratively refine landmarks in a coarse-to-fine manner. 
At each iteration, prototype learning with similarity-based matching is employed to capture local cues and predict robust offsets. 
Finally, an error-aware regularization is introduced to stabilize training and further improve localization performance.
In future work, we plan to extend our method to more organs, modalities, and 3D scenarios.

\begin{credits}
\subsubsection{\ackname} This work was supported by the Frontier Technology Development Program of Jiangsu Province (No. BF2024078) and National Natural Science Foundation of China (No. 12326619).
\subsubsection{\discintname}
The authors have no competing interests to declare that are relevant to the content of this article.
\end{credits}

\bibliographystyle{splncs04}
\bibliography{reference}

\end{document}